\documentclass[journal,transmag]{IEEEtran}

\ifCLASSINFOpdf
\usepackage{graphicx}  

\usepackage{amsmath}
\usepackage{amssymb}

\begin{document}

\title{\textbf{NeuSOGA3D: A Neuro-Symbolic Framework for Explainable 3D Geometric Reconstruction}}

\author{
\IEEEauthorblockN{
Qingde Li\IEEEauthorrefmark{1},
Qingqi Hong\IEEEauthorrefmark{2},
Zihan Li\IEEEauthorrefmark{3},
Jie Tian\IEEEauthorrefmark{4},~\IEEEmembership{Fellow,~IEEE}}
\IEEEauthorblockA{\IEEEauthorrefmark{1}Computer Science, School of Digital and Physical Sciences, University of Hull, HU6 7RX, UK}
\IEEEauthorblockA{\IEEEauthorrefmark{2}The Institute of Artificial Intelligence, Xiamen University, Xiamen, China}
\IEEEauthorblockA{\IEEEauthorrefmark{3}Department of Bioengineering, University of Washington, Seattle, WA, USA}
\IEEEauthorblockA{\IEEEauthorrefmark{4} Chinese Academy of Sciences, Institute of Automation, Beijing, China}
}

\markboth{Preprint. Under review.}%
{Shell \MakeLowercase{\textit{et al.}}: Bare Demo of IEEEtran.cls for IEEE Transactions on xxxx Journals}

\IEEEtitleabstractindextext{%


\begin{abstract} 

Three-dimensional reconstruction from unorganized point clouds remains a fundamental challenge in computer vision, geometric modeling, reverse engineering, and computer-aided design. Recent advances in neural implicit representations, including signed-distance networks, neural radiance fields, and foundation-model-based geometric systems, have demonstrated remarkable reconstruction capabilities. However, these approaches typically encode geometry within high-dimensional latent representations, limiting interpretability, geometric traceability, and direct reuse within downstream engineering workflows.

This paper introduces \textbf{NeuSOGA3D} (\textbf{Neu}ro-\textbf{S}ymbolic \textbf{O}bservation-Guided Geometric \textbf{A}bstraction in \textbf{3D}), an extension of NeuSOGA~\cite{Li2026NeuSOGA} that enables explainable three-dimensional geometric reconstruction from unorganized point clouds. Inspired by theories of human spatial perception, NeuSOGA3D combines observation-guided perceptual abstraction inherited from NeuSOGA with explicit symbolic geometric reasoning, progressively transforming geometric observations into continuous three-dimensional representations while preserving interpretability throughout the reconstruction process.

The framework first performs multi-view geometric abstraction by projecting an unorganized point cloud onto principal orthographic planes and constructing symbolic implicit spline representations from the resulting observations. These view-dependent representations are fused through shape-preserving constructive solid geometry operators to generate a coarse visual-hull hypothesis. To recover geometric structures that cannot be inferred from orthographic observations alone, the framework subsequently performs multi-axial cross-sectional decomposition and volumetric reconstruction using Partial Shape-Preserving Splines (PSPS). The PSPS basis functions are constructed directly from a family of shape-preserving piecewise polynomial transition functions \(H(s,n)\), producing continuous volumetric representations that preserve structural organization, cross-sectional morphology, and topological consistency. Unlike conventional neural implicit approaches, which encode geometry within trainable network parameters, NeuSOGA3D progressively transforms observations into explicit symbolic entities including control polygons, implicit spline fields, cross-sectional representations, PSPS volumetric lofts, and constructive solid geometry operators. Consequently, the resulting model supports both symbolic geometric reasoning and direct translation into CAD-compatible representations such as B-splines, NURBS surfaces, functional representations, and boundary-representation (B-Rep) models.

Experimental evaluation across all forty categories of the ModelNet40 benchmark demonstrates the generality of the proposed framework and its ability to recover structurally meaningful geometric representations from diverse point-cloud observations. The results further demonstrate explainable geometric reconstruction, symbolic abstraction of dense observations into reusable control structures, and direct generation of CAD-ready geometric representations. These findings suggest that geometric understanding can emerge through the interaction of learned perception and symbolic geometric reasoning, and position NeuSOGA3D as a step toward explainable geometric intelligence. \end{abstract}


}

\maketitle

\begin{figure*}[t] \centering \includegraphics[width=\textwidth]{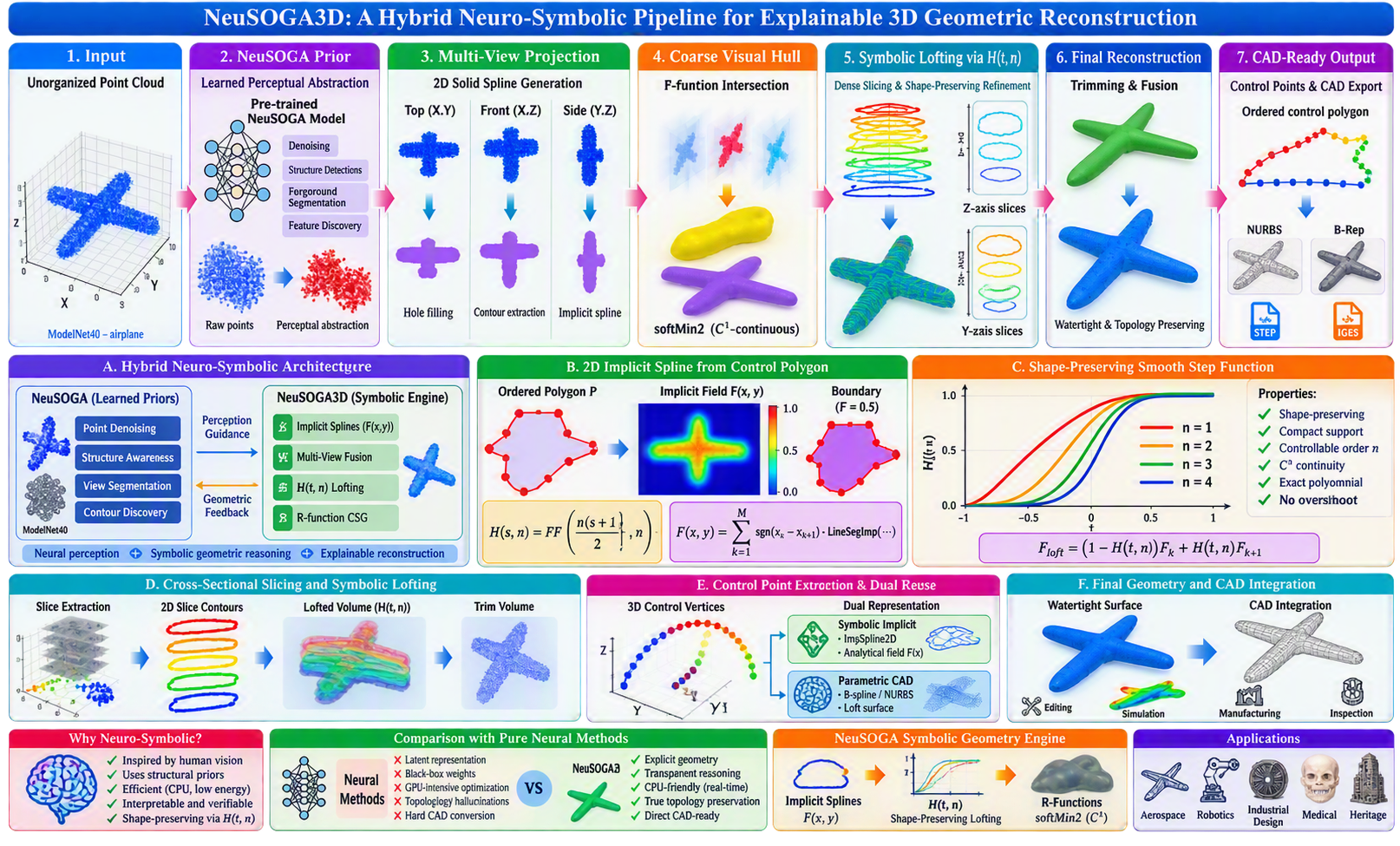} 

\caption{ Architectural overview of the proposed NeuSOGA3D framework. Inspired by the interaction between innate structural priors and acquired perceptual experience in human spatial intelligence, NeuSOGA3D adopts a hybrid neuro-symbolic reconstruction strategy that combines learned perceptual abstraction with explicit symbolic geometric reasoning. Given an unorganized point cloud, the framework first performs multi-view geometric abstraction by projecting the observations onto the principal orthographic planes and constructing watertight NeuSOGA implicit spline representations. These symbolic views are subsequently fused through shape-preserving constructive solid geometry operators to generate a coarse visual-hull hypothesis. To recover geometric structures that cannot be inferred from orthographic observations alone, the framework performs multi-axial cross-sectional decomposition and volumetric reconstruction. Cross-sectional implicit fields extracted along the three principal directions are converted into continuous volumetric representations through Partial Shape-Preserving Splines (PSPS), whose basis functions are generated directly from the smooth piecewise polynomial transition function family $H(s,n)$. The resulting volumetric hypotheses are combined through localized shape-preserving $R$-function blending, producing a continuous, topology-preserving implicit representation while maintaining explicit symbolic control structures throughout the reconstruction pipeline. Unlike neural implicit methods that encode geometry within latent network parameters, NeuSOGA3D progressively transforms observations into interpretable symbolic entities, including contour abstractions, implicit spline fields, PSPS volumetric lofts, constructive solid geometry operators, and reusable control polygons. These symbolic structures provide complete geometric traceability and may be directly reused to generate CAD-compatible representations such as B-splines, NURBS surfaces, boundary representations (B-Reps), and functional solid models. Consequently, NeuSOGA3D reconstructs not only continuous geometry but also an explainable and reusable geometric representation suitable for downstream reasoning, engineering design, and CAD/CAM applications. }
\label{fig:overview} \end{figure*}

\begin{figure*}[t] 
\centering \includegraphics[width=\textwidth]{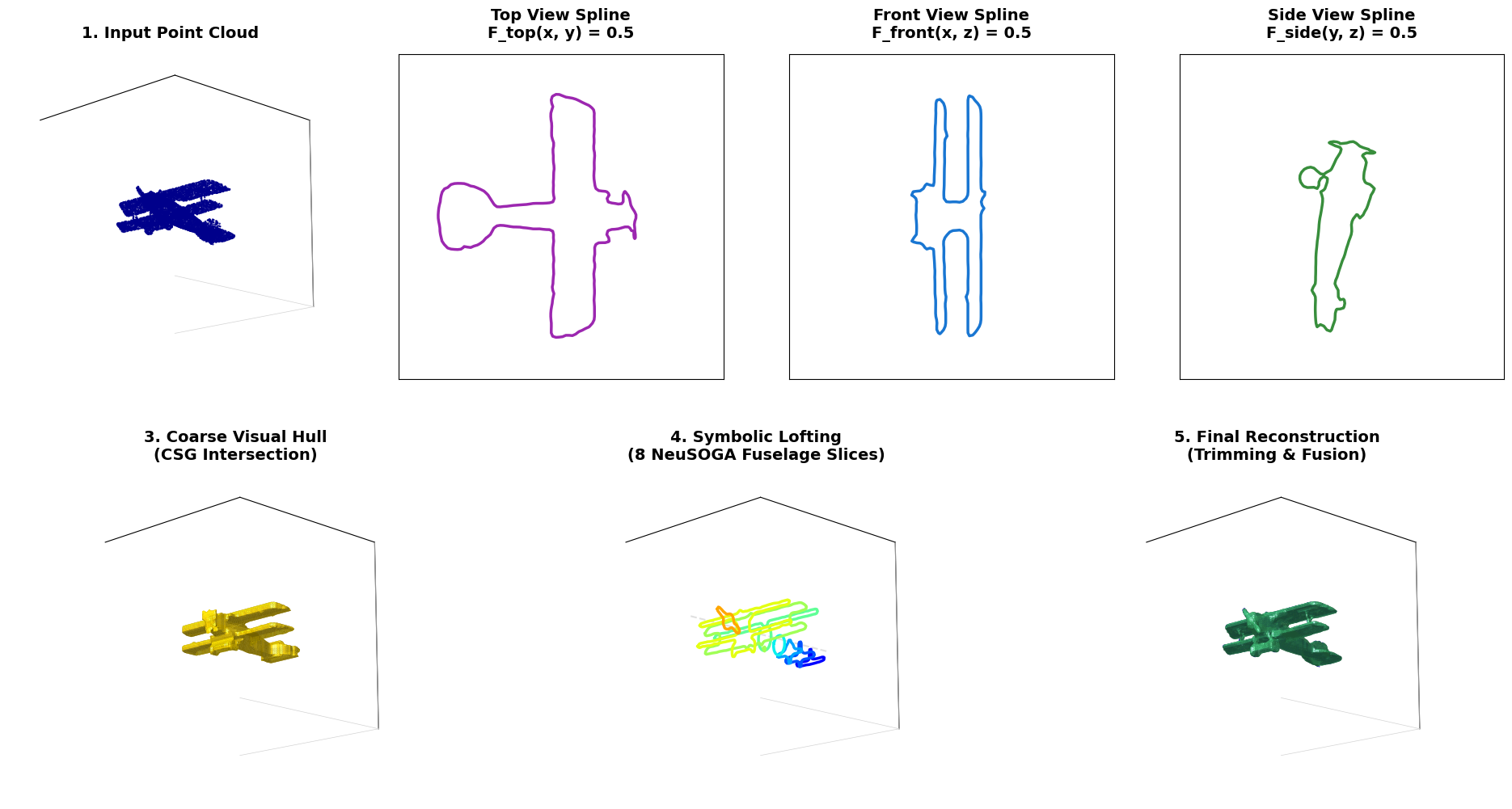} \\
\caption{ Illustration of the NeuSOGA3D reconstruction pipeline on an aircraft point cloud. (1) Input point-cloud observations. (2) Extraction of top, front, and side-view implicit spline contours generated from projected observations. (3) Construction of a coarse visual hull through constructive solid geometry (CSG) intersection of the projected implicit volumes. (4) Symbolic lofting using multiple NeuSOGA-derived fuselage cross-sections to refine the geometric structure. (5) Final watertight reconstruction obtained through trimming, fusion, and implicit surface blending. The example demonstrates how symbolic geometric reasoning progressively transforms sparse point-cloud observations into a smooth, topology-consistent 3D model. }
\label{fig:biplane} \end{figure*}

\IEEEdisplaynontitleabstractindextext

\IEEEpeerreviewmaketitle


\section{Introduction} 

The ability to reconstruct three-dimensional geometry from incomplete observations lies at the core of computer vision, geometric modeling, robotics, reverse engineering, and computer-aided design (CAD). Recent years have witnessed remarkable advances in machine-learning-based reconstruction, particularly through neural implicit representations such as Occupancy Networks, DeepSDF, Neural Radiance Fields (NeRF), VolSDF, NeuS, and Neuralangelo. These methods represent geometry as continuous functions encoded within deep neural networks and have demonstrated impressive capabilities in reconstructing highly complex surfaces from sparse observations.

Despite their success, an important question remains regarding the extent to which latent neural representations support interpretable geometric reasoning and reusable geometric abstraction. Modern neural reconstruction frameworks primarily encode geometric information within high-dimensional latent parameter spaces. Although such representations can produce visually convincing results, the resulting geometric knowledge often lacks direct correspondence to explicit geometric entities such as contours, cross-sections, feature curves, control points, or constructive solid representations. Consequently, geometric information is difficult to inspect, verify, manipulate, or reuse directly within downstream engineering workflows. 

The limitations extend beyond interpretability. Because geometry is represented implicitly through learned parameters, reconstruction quality may depend on training procedures, initialization strategies, model architecture, hyperparameter selection, and data availability. Furthermore, the resulting representations frequently require extensive post-processing before integration into conventional CAD/CAM environments based on explicit boundary representations (B-Reps), spline control structures, and parametric design models. This creates a disconnect between contemporary AI-based reconstruction and engineering-oriented geometric modeling. 

More fundamentally, the prevailing paradigm of contemporary deep learning differs substantially from biological intelligence. The human visual system acquires sophisticated spatial understanding while operating with extremely limited computational resources relative to modern large-scale learning systems. Rather than learning every possible geometric configuration from vast quantities of data, humans appear to combine intrinsic structural priors with accumulated experience to infer the underlying organization of their environment. Classical theories of perception suggest that visual understanding emerges through abstraction, structural reasoning, and hypothesis formation rather than through brute-force optimization alone. 

From an early age, humans acquire an understanding of object permanence, boundaries, continuity, enclosure, and topological relationships. When observing unfamiliar objects, humans do not reconstruct geometry by optimizing millions of parameters. Instead, they exploit powerful geometric priors to identify silhouettes, contours, cross-sections, and structural landmarks, progressively refining an initial global interpretation through local geometric reasoning. This interaction between innate structure and acquired experience enables robust three-dimensional understanding from sparse and incomplete observations. 

Recent developments in neuro-symbolic artificial intelligence suggest that robust intelligence emerges through the interaction between learned perceptual capabilities and explicit symbolic reasoning. Rather than viewing learning and reasoning as competing paradigms, neuro-symbolic systems seek to combine the generalization capabilities of neural models with the transparency, compositionality, and interpretability of symbolic representations. This perspective provides a natural foundation for the development of explainable geometric intelligence. 

The contrast between biological perception and contemporary neural reconstruction motivates the central hypothesis underlying this work: 

\begin{quote} \emph{ Three-dimensional geometric understanding should emerge from interpretable symbolic abstraction and geometric reasoning rather than solely from latent statistical representations. } 
\end{quote} 

Motivated by this hypothesis, we propose \textbf{NeuSOGA3D} (\textbf{Neu}ro-\textbf{S}ymbolic \textbf{O}bservation-Guided Geometric \textbf{A}bstraction in \textbf{3D}), a hybrid neuro-symbolic framework that builds upon the NeuSOGA paradigm~\cite{Li2026NeuSOGA}. NeuSOGA3D extends observation-guided symbolic abstraction from two-dimensional geometric representations to explainable three-dimensional reconstruction. Rather than encoding geometry within latent neural fields, the framework progressively transforms point-cloud observations into interpretable symbolic entities, including contours, implicit spline representations, cross-sectional control structures, volumetric spline models, and constructive solid geometry operators. Consequently, every stage of reconstruction remains geometrically traceable, interpretable, and directly translatable into CAD-compatible representations.

As illustrated in Figure~\ref{fig:overview}, NeuSOGA3D operationalizes a biologically inspired instinctive-to-cognitive reconstruction paradigm. The first stage performs global geometric abstraction through orthographic decomposition, projecting an unorganized point cloud onto multiple principal views and constructing watertight symbolic implicit spline representations. These symbolic observations are fused through shape-preserving constructive solid geometry operators to generate an initial visual-hull hypothesis that captures the global object structure. 

The second stage performs cognitive refinement through multi-axial cross-sectional reasoning. Inspired by the human ability to infer interior structure from successive contour observations, the point cloud is decomposed into dense families of symbolic cross-sections extracted along multiple principal directions. The resulting cross-sectional representations are transformed into continuous volumetric fields through Partial Shape-Preserving Splines (PSPS), whose basis functions are generated directly from the smooth transition function family $H(s,n)$. Independent volumetric hypotheses reconstructed along multiple axes are subsequently fused through shape-preserving constructive solid geometry operators to form a topology-preserving geometric consensus. This refinement stage resolves projection ambiguities, preserves structural organization, and recovers geometric information that cannot be inferred from orthographic observations alone. 

Unlike conventional neural implicit methods, every stage of the proposed pipeline remains analytically defined and geometrically interpretable. Geometric information is represented explicitly through symbolic control structures rather than hidden latent variables. Consequently, the reconstructed geometry maintains direct correspondence to contours, implicit spline fields, cross-sectional representations, constructive solid geometry primitives, and reusable control polygons. These symbolic structures can be readily translated into engineering-ready representations including B-splines, NURBS surfaces, boundary representations, and functional solid models. 

From this perspective, NeuSOGA3D should not be viewed merely as a 3D reconstruction algorithm. Rather, it is a neuro-symbolic framework for geometric intelligence in which learned perception supports symbolic geometric abstraction, reasoning, and representation reuse. 

\subsection{Primary Contributions} 

The primary contributions of this work are summarized as follows: 

\begin{enumerate} 

\item \textbf{Hybrid Neuro-Symbolic Reconstruction Framework.} NeuSOGA3D integrates learned perceptual priors inherited from NeuSOGA with deterministic symbolic geometric reasoning. Neural learning is restricted to perceptual abstraction, while geometric reconstruction is performed through explicit symbolic operators including implicit splines, Partial Shape-Preserving Splines, and shape-preserving constructive solid geometry. 

\item \textbf{Multi-Axial Neuro-Symbolic Geometric Reasoning.} We introduce a reconstruction strategy inspired by the interaction between instinctive global perception and cognitive local reasoning. The framework combines multi-view symbolic abstraction with multi-axial PSPS volumetric reconstruction and shape-preserving geometric consensus. \item \textbf{Explainable Functional Geometric Reconstruction.} The proposed framework generates continuous watertight geometry through explicit symbolic geometric operations and maintains complete traceability between input observations and reconstructed representations. 

\item \textbf{Shape-Preserving Symbolic CSG Formulation.} A localized polynomial $R$-function blending operator is introduced to support smooth constructive solid geometry operations while suppressing the volumetric inflation commonly observed in conventional implicit blending approaches. 

\item \textbf{Direct CAD-Compatible Geometric Abstraction.} NeuSOGA3D preserves explicit geometric control structures throughout reconstruction, enabling direct extraction of reusable symbolic entities suitable for B-splines, NURBS surfaces, boundary representations, functional representations, and broader CAD/CAM workflows without reverse engineering. 
\end{enumerate}


\section{Related Work} 

\subsection{From Statistical Learning to Geometric Intelligence} 

The remarkable success of deep learning has established data-driven optimization as the dominant paradigm in artificial intelligence. Modern foundation models, large language models, and neural reconstruction systems achieve impressive performance by learning high-dimensional statistical representations from massive datasets. Within three-dimensional vision, this philosophy has led to the emergence of neural implicit representations, occupancy fields, signed distance functions, and neural radiance fields, which encode geometry within large collections of trainable parameters \cite{Mescheder2019OccNet,Park2019DeepSDF,Mildenhall2020NeRF,Wang2021NeuS}. Despite their effectiveness, growing concerns have emerged regarding the interpretability and scientific transparency of deep neural networks. Numerous studies in Explainable Artificial Intelligence (XAI) have highlighted that neural models frequently operate as black-box systems in which internal representations possess limited semantic correspondence to the underlying reasoning process \cite{DoshiVelez2017,Rudin2019,Lipton2018}. As model complexity increases, understanding why a particular prediction or reconstruction is generated becomes increasingly difficult. This challenge is particularly significant in scientific and engineering applications, where reproducibility, verifiability, and analytical guarantees are often as important as predictive accuracy. The limitations of purely data-driven learning become especially apparent in geometric reconstruction. While neural networks excel at approximating surfaces, they generally do not provide explicit geometric abstractions such as boundaries, contours, cross-sections, constructive solids, or parametric control structures. Consequently, geometry is represented implicitly through numerical parameters rather than through interpretable geometric entities. Recent discussions in geometric artificial intelligence have therefore emphasized the need for models that combine learning with explicit structural reasoning \cite{Battaglia2018,Marcus2020,GarneloShanahan2019}.

\subsection{Interpretability and Explainability in 3D Vision} 

The growing adoption of deep neural networks for three-dimensional reconstruction and scene understanding has intensified concerns regarding transparency, interpretability, and trustworthiness. Neural implicit representations such as DeepSDF \cite{Park2019DeepSDF}, Occupancy Networks \cite{Mescheder2019OccNet}, NeRF \cite{Mildenhall2020NeRF}, NeuS \cite{Wang2021NeuS}, and Neuralangelo \cite{Li2023Neuralangelo} achieve impressive reconstruction quality by representing geometry within high-dimensional parameter spaces. However, the relationship between these learned representations and explicit geometric entities remains largely opaque. 

Recent surveys of neural radiance fields and implicit neural representations have highlighted the challenges associated with interpreting learned volumetric fields and verifying the geometric processes that produce reconstructed surfaces \cite{Gao2025NeRFReview}. While such methods provide powerful function approximators, geometry is generally encoded through latent numerical representations rather than explicit geometric abstractions such as contours, cross-sections, feature curves, constructive solids, or control structures. 

The broader explainable artificial intelligence (XAI) community has similarly identified transparency, traceability, and interpretability as important requirements for the deployment of AI systems in scientific and engineering applications \cite{DoshiVelez2017,Rudin2019,Lipton2018}. These concerns are particularly relevant for geometric reconstruction, where the ability to inspect, validate, modify, and reuse reconstructed geometry is often as important as numerical reconstruction accuracy. 

NeuSOGA3D addresses this problem from a fundamentally different perspective. Rather than attempting to explain latent neural representations after reconstruction, the framework represents geometry explicitly through symbolic entities throughout the entire reconstruction process. Consequently, explainability is embedded directly into the geometric representation rather than imposed as a post-hoc analysis of a learned model.

\subsection{Biological Vision, Structural Priors and Neuro-Symbolic Intelligence} 

Human intelligence appears to operate according to principles that differ fundamentally from large-scale statistical optimization. Contemporary theories of perception, including ecological perception, predictive processing, and active inference \cite{Friston2013}, suggest that perception emerges through the interaction between innate structural priors and sensory observations. 

Rather than learning all geometric regularities from scratch, biological vision exploits assumptions regarding boundaries, continuity, symmetry, enclosure, and topological organization to infer spatial structure from incomplete observations. Human observers routinely recover volumetric structure, cross-sectional organization, and occluded geometry from sparse visual information while operating with extremely limited computational resources compared with modern AI systems. 

The importance of structural priors and relational reasoning has also been emphasized within machine learning. Battaglia et al. \cite{Battaglia2018RelationalInductiveBiases} argued that relational inductive biases provide an important foundation for learning efficient representations from limited observations. Such findings have motivated increasing interest in neuro-symbolic artificial intelligence, which seeks to combine the perceptual capabilities of neural systems with the transparency and compositionality of symbolic reasoning \cite{Garcez2019,Garcez2022NeuralSymbolicSurvey}. 

Although neuro-symbolic systems have demonstrated promising results in logical reasoning, concept learning, and visual understanding, their application to geometric reconstruction remains limited. Most existing frameworks focus primarily on semantic interpretation rather than explicit geometric representation and reconstruction.

\subsection{Neuro-Symbolic Geometric Modeling} While neuro-symbolic approaches have traditionally focused on language, logic, and scene understanding, recent research has begun extending symbolic reasoning into geometric domains. Early frameworks demonstrated how symbolic representations could support interpretable spatial reasoning \cite{Yi2018NeuralSymbolicVQA}, while more recent systems have incorporated structured intermediate representations into 3D scene understanding and vision-language models \cite{Hsu2023NS3D,Garcez2022NeuralSymbolicSurvey,Mo2026APEIRIA}. Nevertheless, these methods generally employ symbolic abstractions to explain the behaviour of underlying neural networks rather than replacing the reconstruction process itself. Geometry remains encoded within learned latent representations, and symbolic components are often introduced only at higher semantic levels. In contrast, NeuSOGA3D adopts a stronger neuro-symbolic philosophy in which the geometric reconstruction process itself is symbolic, deterministic, and analytically interpretable. Every stage of the pipeline is expressed through explicit geometric operations involving implicit splines, cross-sectional abstractions, and constructive solid geometry rather than through latent neural embeddings.

\subsection{Slice-Based Reconstruction and Contour Lofting} 

Cross-sectional reconstruction has played a central role in computer-aided geometric design, medical image processing, and reverse engineering. Early contour-based methods reconstructed three-dimensional objects by interpolating between planar sections using spline and lofting techniques \cite{Piegl1997NURBS,Farin2002Curves}. Similar methodologies have been widely employed in medical imaging, where anatomical structures are reconstructed from sparse contour observations extracted from tomographic data \cite{Bajaj1996,Boissonnat1988}. 

More recent research has revisited slice-based reconstruction for point clouds and image observations. Methods based on contour interpolation, section-aware meshing, and cross-sectional feature extraction have demonstrated improved robustness when reconstructing geometrically complex and topologically challenging objects. 

Recent developments in shape-preserving spline theory have further demonstrated that smooth transition functions can be transformed into non-negative partition-of-unity basis functions suitable for robust geometric interpolation and lofting. In particular, the Partial Shape-Preserving Spline (PSPS) framework \cite{Li2011PSPS} provides an analytical alternative to conventional Hermite and B-spline formulations, offering compact support, controllable smoothness, locality, and strong shape-preserving behaviour. 

NeuSOGA3D follows a distinctly symbolic geometric paradigm. Instead of learning a volumetric representation globally, the framework performs dense multi-axial decomposition of the point cloud, reconstructs each section as a NeuSOGA implicit spline, and subsequently generates continuous volumetric fields through Partial Shape-Preserving Spline (PSPS) reconstruction. The resulting volumetric hypotheses are fused through shape-preserving constructive solid geometry operators to form a geometric consensus representation. This process preserves cross-sectional morphology, structural organization, and topological consistency while maintaining a fully analytical and interpretable geometric representation.

\subsection{CAD Reconstruction and Boundary Representation Generation} The increasing demand for editable engineering models has motivated substantial research into converting point clouds, meshes, and images into CAD and Boundary Representation (B-Rep) models. The introduction of the ABC dataset \cite{Koch2019ABC} provided a large-scale benchmark containing approximately one million CAD models and significantly accelerated data-driven CAD research. Early learning-based approaches such as DeepCAD \cite{Wu2021DeepCAD} demonstrated the feasibility of generating CAD construction sequences using transformer architectures. Subsequently, methods including BRepNet \cite{Lambourne2021BRepNet}, Point2CAD \cite{Vietri2022Point2CAD}, Free2CAD \cite{Li2022Free2CAD}, CC3D-Ops \cite{Dupont2022CC3DOps}, CADParser \cite{Zhou2023CADParser}, and BrepGen \cite{Xu2023BrepGen} focused on recovering engineering representations from geometric observations. The last two years have witnessed rapid growth in foundation-model-driven CAD systems. Representative examples include Text2CAD \cite{Khan2024Text2CAD}, Query2CAD\cite{Xie2025TextToCadQuery}, CAD-Llama\cite{Li2025CADLlama}, Img2CAD \cite{Chen2025Img2CAD}, OpenECAD \cite{Yuan2024OpenECAD}, OmniCAD \cite{Wang2026OmniCAD}, CAD-Recode \cite{Rukhovich2025CADReCode}, LLM4CAD \cite{Li2025LLM4CAD}, and Text-to-CadQuery \cite{Xie2025TextToCadQuery}. These systems leverage large language models and multimodal reasoning to infer design intent, procedural histories, and parametric operations directly from textual and geometric descriptions. While highly scalable, most CAD-generation frameworks remain probabilistic and may produce geometrically inconsistent design histories. NeuSOGA3D adopts a fundamentally different philosophy by extracting geometry directly from observations. The resulting representation is deterministic, geometrically faithful, and naturally compatible with NURBS and CAD-based workflows. 

\subsection{Neuro-Symbolic 3D Reasoning and Explainable Spatial Intelligence} 

Neuro-symbolic artificial intelligence seeks to combine the pattern-recognition capabilities of neural networks with the transparency, compositionality, and reasoning capabilities of symbolic systems \cite{Garcez2022NeuralSymbolicSurvey,Garcez2019}. Early neuro-symbolic architectures demonstrated that explicit symbolic representations could substantially improve interpretability and generalization in visual reasoning tasks \cite{Yi2018NeuralSymbolicVQA}. 

Recent developments have extended these ideas to three-dimensional reasoning and spatial intelligence. NS3D \cite{Hsu2023NS3D} introduced neuro-symbolic grounding of three-dimensional objects and spatial relationships through explicit concept representations. More recently, systems such as SceneCOT \cite{Linghu2026SceneCOT}, Lexicon3D \cite{Man2024Lexicon3D}, and APEIRIA \cite{Mo2026APEIRIA} have explored interpretable reasoning processes through symbolic intermediate representations and explainable reasoning traces. 

Despite these advances, most neuro-symbolic systems focus primarily on semantic scene understanding, question answering, or embodied reasoning. The underlying geometric representation typically remains encoded within neural latent spaces. In contrast, NeuSOGA3D extends neuro-symbolic principles directly to geometric reconstruction itself. Geometry is represented through explicit symbolic splines, PSPS volumetric fields, constructive solid geometry operators, and reusable control structures, making interpretability an intrinsic property of the reconstruction process rather than a post-hoc explanation.

\subsection{Constructive Solid Geometry, Functional Representations and R-Functions} Functional geometric representations (F-Reps) provide an alternative to explicit boundary-based modeling by describing solid objects through continuous mathematical functions \cite{Pasko1995FRep}. Rooted in the theory of R-functions introduced by Rvachev \cite{Rvachev1963}, F-Reps enable constructive solid geometry (CSG) operations to be performed directly on implicit functions. Subsequent developments by Shapiro and co-workers established a rigorous mathematical framework for representing and manipulating solids using real-valued functions \cite{Shapiro1994,Shapiro2007}. While classical R-function formulations provide powerful tools for smooth Boolean operations, many blending operators exhibit asymptotic influence and may introduce global geometric distortion, volumetric swelling, or loss of local shape characteristics. To address these issues, shape-preserving blending methodologies have been investigated for implicit geometric modeling. In particular, Li \cite{Li2007Blending} introduced a family of smooth piecewise polynomial blending operations capable of achieving local support, controllable smoothness, and shape-preserving behaviour for implicit objects. These properties later influenced the development of the shape-preserving transition functions \cite{Li2007Transition}, Partial Shape-Preserving Splines (PSPS) \cite{Li2011PSPS}, and the localized blending operators adopted in NeuSOGA3D. The proposed framework extends this line of research by integrating shape-preserving symbolic blending with implicit spline representations and multi-axial PSPS volumetric lofting. Unlike conventional neural implicit fields, which encode geometry through learned parameters, NeuSOGA3D performs volumetric construction and refinement through explicit shape-preserving symbolic operators, thereby preserving both geometric interpretability and topological consistency.

\subsection{Research Gap} 

Despite significant advances in neural reconstruction, CAD generation, neuro-symbolic reasoning, and functional geometric modeling, a fundamental gap remains between geometric approximation and geometric understanding. 

Neural reconstruction methods achieve impressive reconstruction quality but generally represent geometry through latent parameter spaces that are difficult to interpret, verify, edit, or reuse directly. Conversely, classical geometric modeling frameworks provide mathematically rigorous and interpretable representations but typically lack mechanisms for robust reconstruction from unstructured observations. 

Similarly, recent neuro-symbolic systems have demonstrated the value of combining learning and symbolic reasoning, yet their symbolic components are predominantly applied at semantic and relational levels rather than within the geometric reconstruction process itself. 

To the best of our knowledge, existing approaches do not provide a unified framework that simultaneously supports symbolic geometric abstraction, explainable reconstruction, topology-preserving volumetric modeling, and direct generation of CAD-compatible representations from point-cloud observations. 

NeuSOGA3D addresses this gap through a hybrid neuro-symbolic architecture that combines learned perceptual abstraction with symbolic implicit splines, Partial Shape-Preserving Splines (PSPS), and shape-preserving constructive solid geometry. The resulting framework reconstructs not only continuous geometry but also reusable symbolic geometric representations suitable for geometric reasoning, engineering design, and CAD/CAM applications.


\section{Theoretical Foundations of NeuSOGA3D} 

Unlike conventional neural implicit reconstruction methods, NeuSOGA3D is neither a purely data-driven model nor a purely symbolic geometric system. Instead, it adopts a hybrid neuro-symbolic architecture in which learned perceptual priors are combined with explicit symbolic geometric reasoning. 

The framework builds upon NeuSOGA, whose underlying representation incorporates pre-trained perception components that provide robust geometric abstraction from observed data. However, unlike neural implicit methods where geometry is encoded directly within trainable network parameters, NeuSOGA3D progressively transforms geometric observations into explicit symbolic entities including contours, control polygons, implicit spline functions, cross-sectional representations, volumetric spline fields, and constructive solid geometry (CSG) operators. 

Consequently, machine learning serves only as a perceptual abstraction mechanism. Geometric reconstruction itself is performed through deterministic symbolic operators that preserve complete geometric traceability. This design follows the central principle of neuro-symbolic intelligence by combining the generalization capabilities of learned priors with the transparency, interpretability, and verifiability of symbolic reasoning. 

The symbolic geometry engine of NeuSOGA3D is built upon three mathematically connected developments introduced previously for symbolic geometric modelling: 

\begin{enumerate} 

\item \textbf{Shape-Preserving Piecewise Polynomial Transition Functions} \cite{Li2007Transition}; 

\item \textbf{Partial Shape-Preserving Splines (PSPS)} \cite{Li2011PSPS}; 

\item \textbf{Algebraic Implicit Splines} \cite{Li2009ImplicitSpline}. 
\end{enumerate} 

NeuSOGA3D extends these developments into a unified neuro-symbolic framework for volumetric geometric reconstruction and constructive solid geometry. 

\subsection{Neural Priors and Symbolic Reconstruction} 

NeuSOGA3D adopts the neuro-symbolic hypothesis that efficient spatial understanding emerges from the interaction between learned perceptual priors and explicit structural reasoning. This viewpoint is consistent with contemporary theories of biological perception, which suggest that humans do not reconstruct geometry solely through data accumulation but instead combine prior knowledge with symbolic spatial abstractions. 

The only learned component in NeuSOGA3D is the observation module inherited from NeuSOGA, which leverages a pre-trained Segment Anything Model (SAM) to derive geometric priors from raw point-cloud data. Once these priors have been extracted, the reconstruction process proceeds entirely through symbolic geometric reasoning and procedural surface generation.

Unlike neural implicit representations, no geometric surface is encoded within latent vectors, embeddings, or trainable neural weights. Instead, geometric information is represented through explicit polygonal contours, implicit spline functions, volumetric spline fields, and constructive solid geometry operators. 

This separation between perception and reconstruction forms the central neuro-symbolic principle of NeuSOGA3D. Neural learning contributes robustness and generalization, whereas geometric reasoning remains analytical, interpretable, and mathematically verifiable. 

\subsection{Shape-Preserving Transition Functions} 

A central mathematical foundation of NeuSOGA3D is the family of smooth piecewise polynomial transition functions introduced in \cite{Li2007Transition}. Unlike conventional interpolation kernels such as Hermite smoothstep functions, these transition functions provide exact compactly-supported polynomial representations with controllable smoothness order and strong shape-preserving behaviour. 

The construction begins with the recursively defined divided-difference generator 

\begin{equation} 
FF(s,n) = \frac{s}{n}FF(s,n-1) + \left( 1-\frac{s}{n} \right) FF(s-1,n-1), 
\end{equation} 

with boundary condition 

\begin{equation} 
FF(s,0) = \begin{cases} 1, & s>0,\\ 0.5, & s=0,\\ 0, & s<0. \end{cases} 
\end{equation} 

The corresponding transition function is defined as 

\begin{equation} 
H(s,n) = FF \left( \frac{n(s+1)}{2}, n \right). 
\end{equation} 

The resulting function family possesses several desirable properties: 

\begin{itemize} 

\item exact piecewise polynomial representation; 

\item compact support; 

\item controllable smoothness order; 

\item monotonic shape-preserving behaviour; 

\item absence of oscillation and overshoot; 

\item straightforward analytical differentiation. 

\end{itemize} 

The transition function \(H(s,n)\) serves as the generating mechanism for both the algebraic implicit spline representation \cite{Li2009ImplicitSpline} and the PSPS framework \cite{Li2011PSPS}. Consequently, a single mathematical primitive underlies contour representation, volumetric lofting, and symbolic reconstruction throughout NeuSOGA3D. 

\subsection{Partial Shape-Preserving Splines} 

Building upon the transition-function framework, the Partial Shape-Preserving Spline (PSPS) methodology \cite{Li2011PSPS} demonstrated that translated copies of \(H(s,n)\) can be transformed into spline-like basis functions satisfying locality, non-negativity, and partition-of-unity constraints. 

Given a family of translated transition functions 

\[ H_i(y), \] 

the corresponding PSPS basis functions are defined as 

\begin{equation} 
B_i(y) = H_{i-1}(y)-H_i(y). 
\end{equation} 

These basis functions satisfy 

\begin{equation} 
B_i(y)\ge 0, 
\end{equation} 
and 
\begin{equation} 
\sum_i B_i(y)=1. 
\end{equation} 

Consequently, the resulting spline basis inherits compact support, locality, numerical stability, and shape-preserving behaviour while maintaining an exact analytical representation. 

Unlike conventional spline formulations based on Cox-de Boor recursion, the PSPS basis is generated directly from the transition function family. Smoothness order, support width, and transition behavior therefore remain explicitly controllable through the underlying symbolic framework. 

\subsection{Symbolic 2D Algebraic Spline Engine} 

The 2D algebraic implicit spline formulation developed by Li and Tian \cite{Li2009ImplicitSpline} provides the symbolic contour representation used throughout NeuSOGA3D. 

Given an ordered control polygon 

\[ \mathcal{P} = \left\{ \mathbf{p}_k=(x_k,y_k) \right\}_{k=1}^{M}, \] 

NeuSOGA constructs a continuous implicit field directly from symbolic boundary segments without solving global interpolation systems or differential equations. 

The resulting implicit field may be expressed in the form
\begin{equation} F(x,y) = \sum_{k=1}^{M} \mathrm{sgn}(x_k-x_{k+1}) \, \mathrm{LineSegImp} \left( x,y, \mathbf{p}_k, \mathbf{p}_{k+1} \right), \label{eq:impspline}
\end{equation} 

where \(\mathrm{LineSegImp}\) denotes the implicit representation associated with each directed boundary segment. 

The shape-preserving property of the proposed implicit spline arises primarily from the polygon-based formulation of Equation~(\ref{eq:impspline}). For sufficiently small non-negative values of $\delta$, the implicit field satisfies $F(x,y)\approx 1$ throughout the interior of the polygon

\[ \mathcal{P}=\left\{\mathbf{p}_k=(x_k,y_k)\right\}_{k=1}^{M}, \]
thereby maintaining the original geometric structure during blending. The transition function $H(s,n)$ plays a complementary role in the local construction of

\[ \mathrm{LineSegImp}(x,y,\mathbf{p}_k,\mathbf{p}_{k+1}), \]
particularly for horizontal and vertical boundary segments, where it contributes to the continuity and locality of the resulting implicit representation.

A notable property of this representation is that the extracted symbolic control structures remain directly reusable within conventional CAD systems. Consequently, the same geometric abstraction simultaneously supports symbolic implicit modelling, B-spline modelling, NURBS generation, and boundary-representation workflows. 

\subsection{Multi-Axial Volumetric Reconstruction Using PSPS} 

The PSPS framework provides the mathematical foundation for transforming symbolic cross-sections into continuous volumetric geometry. 

Let 

\[ \{F_i\}_{i=1}^{K} \] 

denote the implicit slice fields reconstructed from a sequence of extracted cross-sections. The corresponding continuous volumetric field is evaluated as 

\begin{equation} F(\mathbf{x}) = \sum_{i=1}^{K} F_i(\mathbf{x}) B_i(w), \end{equation} 

where \(B_i(w)\) denotes the PSPS basis function associated with the reconstruction direction \(w\). 

This formulation may be interpreted as a volumetric extension of symbolic spline reconstruction in which individual implicit slice fields act as control sections and the PSPS basis functions govern their influence throughout the volume. 

Because the basis functions form a non-negative partition of unity, the resulting field preserves the shape characteristics of the control slices. Thin structures, local extrema, narrow gaps, and cross-sectional morphology therefore remain stable throughout reconstruction. 

Unlike Hermite interpolation, Bezier blending, or globally fitted spline formulations, PSPS reconstruction avoids volumetric swelling, overshoot, clipping of thin structures, and oscillatory behaviour. 

NeuSOGA3D performs this reconstruction independently along all three principal coordinate directions, producing three volumetric hypotheses 

\[ F_X, \qquad F_Y, \qquad F_Z. \] 

Each field represents an independent symbolic interpretation of the observed geometry from a different structural perspective. 

\subsection{Shape-Preserving R-Functions for Constructive Solid Geometry} 

Together with the transition-function framework and PSPS reconstruction, shape-preserving R-functions constitute the third fundamental symbolic operator of the NeuSOGA3D geometry engine. 

The formulation is closely related to earlier work on smooth piecewise polynomial blending operations for implicit geometry \cite{Li2007Blending}, which demonstrated that local shape-preserving blending can avoid the global geometric distortions often introduced by classical asymptotic formulations. 

To combine multiple volumetric hypotheses without generating non-differentiable creases or global geometric distortion, NeuSOGA3D replaces conventional asymptotic R-functions with a compactly supported shape-preserving formulation. 

For two implicit fields \(F_A(\mathbf{x})\) and \(F_B(\mathbf{x})\), define 

\[ d=F_A(\mathbf{x})-F_B(\mathbf{x}), \] 

and 

\[ \tilde d=\frac{2d}{a}, \] 

where \(a>0\) denotes the blending radius. 

The compactly-supported soft absolute value function is defined as 

\begin{equation} S_{abs}(d,a) = \frac{a}{2} \begin{cases} \dfrac{1}{2}\tilde d^{2} \left( 1-\dfrac{|\tilde d|}{6} \right) +\dfrac{2}{3}, & |\tilde d|<2, \\ |\tilde d|, & \text{otherwise}. \end{cases} \end{equation} Equivalently, \begin{equation} S_{abs}(d,a) = \begin{cases} \dfrac{d^2}{a} - \dfrac{|d|^3}{3a^2} + \dfrac{a}{3}, & |d|<a, \\ |d|, & \text{otherwise}. \end{cases} \end{equation} 

The corresponding shape-preserving smooth minimum operator is 

\begin{equation} \mathrm{softMin}_2(F_A,F_B,a) = \frac{1}{2} \Big( F_A+F_B-S_{abs}(F_A-F_B,a) \Big). \end{equation} 

The final NeuSOGA3D reconstruction is obtained through constructive solid geometry consensus: 

\begin{equation} F_{final} = \mathrm{softMin}_2 \Big( F_X, \mathrm{softMin}_2(F_Y,F_Z,a), a \Big). \end{equation} 

This formulation preserves local shape characteristics while suppressing the volumetric inflation commonly associated with classical implicit blending operators. 

\subsection{Summary} From a mathematical perspective, NeuSOGA3D may be viewed as the three-dimensional synthesis of three previously independent developments: smooth piecewise polynomial transition functions \cite{Li2007Transition}, Partial Shape-Preserving Splines \cite{Li2011PSPS}, and algebraic implicit splines \cite{Li2009ImplicitSpline}. By integrating these foundations within a hybrid neuro-symbolic architecture and combining them through shape-preserving constructive solid geometry, NeuSOGA3D extends symbolic geometric modelling from planar contour representation to explainable volumetric geometric reconstruction. 

The resulting framework reconstructs not only continuous geometry but also explicit symbolic geometric representations that remain interpretable, reusable, CAD-compatible, and suitable for geometric reasoning.

\section{Methodology: Neuro-Symbolic Multi-Axial Reconstruction} 

NeuSOGA3D reconstructs geometry through a neuro-symbolic coarse-to-fine reasoning process inspired by the interaction between instinctive perception and cognitive refinement in biological vision. The framework combines learned perceptual priors inherited from NeuSOGA with explicit symbolic geometric reasoning. Rather than encoding geometry within latent neural representations, NeuSOGA3D progressively transforms point-cloud observations into symbolic contour representations, implicit spline fields, volumetric spline abstractions, and constructive solid geometry operators. The reconstruction process consists of two complementary stages. Phase I performs rapid multi-view geometric abstraction that generates a coarse symbolic visual-hull hypothesis. Phase II refines this initial hypothesis through multi-axial cross-sectional reasoning using Partial Shape-Preserving Splines (PSPS). Independent volumetric hypotheses generated from multiple structural perspectives are subsequently integrated through symbolic geometric consensus to produce the final reconstruction. 

\subsection{Phase I: Instinctive Multi-View Geometric Abstraction} 

Given an unorganized point cloud \begin{equation} \mathcal{Q} = \left\{ \mathbf{q}_i=(x_i,y_i,z_i) \right\}_{i=1}^{N}, \end{equation} the observations are first normalized into a reconstruction domain \begin{equation} \Omega \subset \mathbb{R}^{3}. \end{equation} This stage emulates the instinctive component of human visual perception, where a global interpretation of object structure is rapidly inferred from a small number of orthographic observations. 

\paragraph{Orthographic Projection} The point cloud is projected onto the three principal planes \begin{equation} \Pi_{XY}, \qquad \Pi_{XZ}, \qquad \Pi_{YZ}, \end{equation} generating top, front, and side geometric observations. 

\paragraph{Solid Area Recovery} Since point clouds typically sample object boundaries rather than interiors, the resulting projections form sparse contour-like structures. Morphological closing followed by topological hole filling transforms these observations into watertight solid regions \begin{equation} \mathcal{M}_{solid} = \mathrm{BinaryFillHoles} \Big( \mathrm{MorphClose} ( \mathcal{M}_{proj}, \mathcal{K} ) \Big). \end{equation} 

\paragraph{NeuSOGA Implicit Spline Construction} Each recovered region is approximated by an ordered symbolic control polygon \begin{equation} \mathcal{P}_{top}, \qquad \mathcal{P}_{front}, \qquad \mathcal{P}_{side}. \end{equation} Using the NeuSOGA symbolic spline engine, these contours are transformed into continuous implicit fields \begin{equation} F_{top}(x,y), \qquad F_{front}(x,z), \qquad F_{side}(y,z). \end{equation} These fields provide explicit symbolic descriptions of the projected geometry. 

\paragraph{Coarse Visual-Hull Consensus} The three symbolic views are fused through the shape-preserving constructive solid geometry operator \begin{equation} \mathrm{softMin}_{2}, \end{equation} yielding an initial visual-hull hypothesis \begin{equation} F_{proto} = \mathrm{softMin}_{2} \Big( \mathrm{softMin}_{2} ( F_{top}, F_{front}, \varepsilon ), F_{side}, \varepsilon \Big). \end{equation} The resulting field captures the dominant volumetric organization and overall topology of the observed object. 

\subsection{Phase II: Multi-Axial PSPS Refinement and Geometric Consensus} 

Although the visual hull captures the global object structure, it cannot recover geometric information that remains ambiguous under a limited set of projections. Concavities, hidden cavities, thin structures, layered components, and internal separations often remain unresolved. The coarse visual hull \[ F_{proto} \] therefore serves as an initial geometric hypothesis rather than a final reconstruction. The objective of Phase II is to refine and validate this hypothesis through dense cross-sectional symbolic reasoning. To achieve this goal, NeuSOGA3D performs multi-axial reconstruction using Partial Shape-Preserving Splines (PSPS). Rather than privileging a single slicing direction, symbolic volumetric reconstruction is performed independently along all three principal axes. 

\paragraph{Multi-Axial Cross-Sectional Analysis} For each axis \begin{equation} \alpha \in \{x,y,z\}, \end{equation} the point cloud is partitioned into a family of overlapping cross-sectional slices \begin{equation} \mathcal{Q}_{k}^{(\alpha)} = \left\{ \mathbf{q} \in \mathcal{Q} \;\middle|\; |\alpha-\alpha_k| \le \gamma\Delta\alpha \right\}, \end{equation} where $\Delta\alpha$ denotes the slice spacing and $\gamma$ controls overlap. Each slice is converted into a symbolic NeuSOGA implicit contour representation \begin{equation} F_k^{(\alpha)}. \end{equation} This transformation converts sparse three-dimensional observations into structured symbolic geometric sections. Importantly, every cross-section remains represented through explicit contours and symbolic control structures. Unlike latent neural representations, these geometric entities remain directly accessible throughout reconstruction, thereby preserving complete geometric traceability. 

\paragraph{PSPS Volumetric Reconstruction} Following the PSPS formulation introduced in Section III, volumetric reconstruction is performed using basis functions generated directly from the transition function family \(H(s,n)\): \begin{equation} B_i(w) = H_{i-1}(w)-H_i(w). \end{equation} The extracted slice fields act as symbolic control sections within the volumetric spline representation. For each slicing direction, a continuous volumetric field is evaluated as \begin{equation} F_X = \sum_i F_i^{(x)} B_i(x), \end{equation} \begin{equation} F_Y = \sum_i F_i^{(y)} B_i(y), \end{equation} \begin{equation} F_Z = \sum_i F_i^{(z)} B_i(z). \end{equation} Unlike conventional interpolation schemes, PSPS reconstruction preserves local cross-sectional morphology while avoiding oscillation, volumetric swelling, clipping of thin structures, and distortion of local extrema. Each volumetric field therefore represents an independent symbolic geometric interpretation of the observed object from a particular structural perspective. 

\paragraph{Geometric Consensus Through Shape-Preserving CSG} The three reconstructed volumetric fields contain complementary geometric information because different slicing directions expose different structural characteristics. Each field may therefore be interpreted as an independent geometric hypothesis derived from a distinct family of cross-sectional observations. Rather than selecting a preferred direction, NeuSOGA3D integrates these competing hypotheses through symbolic geometric consensus. The first fusion stage constructs \begin{equation} F_{pass} = \mathrm{softMin}_{2} ( F_X, F_Y, \varepsilon ), \end{equation} followed by \begin{equation} F_{refined} = \mathrm{softMin}_{2} ( F_{pass}, F_Z, \varepsilon ). \end{equation} This process mirrors biological spatial perception, where coherent three-dimensional understanding emerges through the integration of multiple partial observations rather than a single geometric viewpoint. The resulting field \begin{equation} F_{refined} \end{equation} constitutes the final NeuSOGA3D reconstruction. The resulting geometry remains continuous, watertight, topology-preserving, and fully traceable to explicit symbolic control structures extracted from the original observations. More importantly, NeuSOGA3D reconstructs not merely a surface but an explicit symbolic geometric representation. The reconstructed implicit fields, cross-sectional abstractions, and control structures remain directly reusable for constructive solid geometry, symbolic reasoning, spline-based representations, and downstream CAD/CAM workflows. From this perspective, reconstruction becomes a process of symbolic geometric abstraction and representation learning rather than geometric surface approximation alone.

\section{Symbolic Geometric Abstraction and Representation Reuse} 

A fundamental distinction between NeuSOGA3D and contemporary neural reconstruction methods lies in how geometric knowledge is represented. Neural implicit approaches encode geometry within distributed network parameters and latent feature spaces. Although such models can reproduce complex surfaces with impressive accuracy, the resulting geometric knowledge is often difficult to interpret, verify, modify, or reuse directly. 

NeuSOGA3D adopts a fundamentally different philosophy. Rather than storing geometry within latent numerical representations, the framework progressively transforms point-cloud observations into explicit symbolic geometric structures through contour abstraction, cross-sectional reasoning, implicit spline construction, and control-structure extraction. Consequently, geometric information remains represented through identifiable geometric entities throughout the reconstruction process. 

As illustrated in Figure~\ref{fig:overview}, the extracted symbolic structures form a common geometric language linking reconstruction, geometric reasoning, and CAD modelling. Thousands of unorganized surface samples are transformed into compact collections of contours, control polygons, spline representations, and constructive solid geometry primitives that explicitly encode object boundaries, cross-sectional morphology, structural transitions, and topological organization.

\subsection{Symbolic Geometric Abstraction} 

The primary objective of NeuSOGA3D is not merely to approximate a geometric surface but to recover a symbolic geometric representation of the observed object. During reconstruction, each cross-sectional observation is represented by an ordered contour \[ \mathcal{P}_k = \left\{ \mathbf{p}_j^{(k)} \right\}_{j=1}^{M_k}, \] 

which serves as an explicit symbolic description of the corresponding geometric section. 

The collection of extracted contours and control polygons provides a compact abstraction of the observed geometry. Rather than preserving geometric information through large collections of distributed numerical parameters, NeuSOGA3D represents geometry through a relatively small set of interpretable structural primitives. 

From a geometric perspective, this process may be viewed as a form of symbolic compression in which dense point-cloud observations are converted into reusable geometric knowledge. The resulting representation preserves the essential shape characteristics of the object while substantially reducing representational complexity. 

This abstraction process mirrors the role of structural reasoning in biological perception, where complex sensory observations are believed to be compressed into compact representations that support efficient reasoning, manipulation, and reuse.

\subsection{Geometric Explainability} 

A direct consequence of symbolic abstraction is geometric explainability. 

Unlike neural reconstruction systems, where geometric information is distributed across latent parameter spaces, every reconstructed feature in NeuSOGA3D can be traced directly to explicit symbolic entities derived from the original observations. Contours, cross-sections, implicit spline fields, PSPS volumetric representations, and control structures remain accessible throughout the reconstruction process. 

Consequently, reconstructed geometry possesses complete geometric provenance. Local modifications correspond directly to modifications of explicit control structures rather than adjustments to opaque latent embeddings. The resulting representation therefore remains fully interpretable, verifiable, and analytically traceable. 

From a neuro-symbolic perspective, the extracted symbolic structures may be interpreted as explicit geometric hypotheses describing the observed object. Geometry is therefore reconstructed through structural reasoning rather than recovered solely through function approximation.

\subsection{Unified Representation Reuse} 

A particularly important consequence of symbolic geometric abstraction is representation reuse. 

Because the extracted control structures remain explicit throughout reconstruction, they can be reused across multiple geometric modelling paradigms without retraining, remeshing, or reverse engineering. Within the NeuSOGA framework, the control structures directly support symbolic implicit modelling, constructive solid geometry operations, and functional representations. 

At the same time, the identical control structures can serve as geometric control entities for conventional engineering representations, including B-splines, NURBS surfaces, and boundary-representation (B-Rep) models. 

This dual representation capability establishes a direct bridge between explainable geometric reasoning and practical CAD modelling. Rather than maintaining separate reconstruction and engineering representations, NeuSOGA3D provides a common symbolic abstraction layer from which multiple downstream descriptions can be generated while preserving consistency, interpretability, and geometric fidelity. 

Consequently, NeuSOGA3D does not merely reconstruct geometry. It reconstructs a reusable geometric representation that supports subsequent reasoning, editing, engineering design, and CAD/CAM workflows. The ability to preserve and reuse explicit symbolic control structures therefore represents one of the principal advantages of the proposed neuro-symbolic framework.

\section{Experimental Evaluation} 

\subsection{Evaluation Philosophy} The objective of NeuSOGA3D is fundamentally different from that of conventional neural reconstruction systems. Existing benchmarks primarily evaluate the accuracy of surface approximation through numerical similarity measures. While such metrics are useful for comparing geometric approximations, they provide limited insight into whether a method has successfully recovered an interpretable geometric representation of the observed object. 

NeuSOGA3D is motivated by a different hypothesis. Inspired by human spatial intelligence, the framework seeks to reconstruct symbolic geometric abstractions from observed data rather than encode geometry within latent numerical representations. Consequently, the central evaluation criteria of this work are structural preservation, explainability, symbolic abstraction, and representation reuse. 

The purpose of the experimental study is therefore not merely to assess how closely a surface approximates the input observations, but to investigate whether the proposed neuro-symbolic framework can recover a reusable and interpretable geometric representation capable of supporting subsequent geometric reasoning and modelling operations.

\subsection{Generality Across ModelNet40} To evaluate the robustness and generality of the proposed framework, NeuSOGA3D was applied to representative instances from all forty categories of the ModelNet40 benchmark. 

\begin{figure*}[t] 
\centering \includegraphics[width=\textwidth]{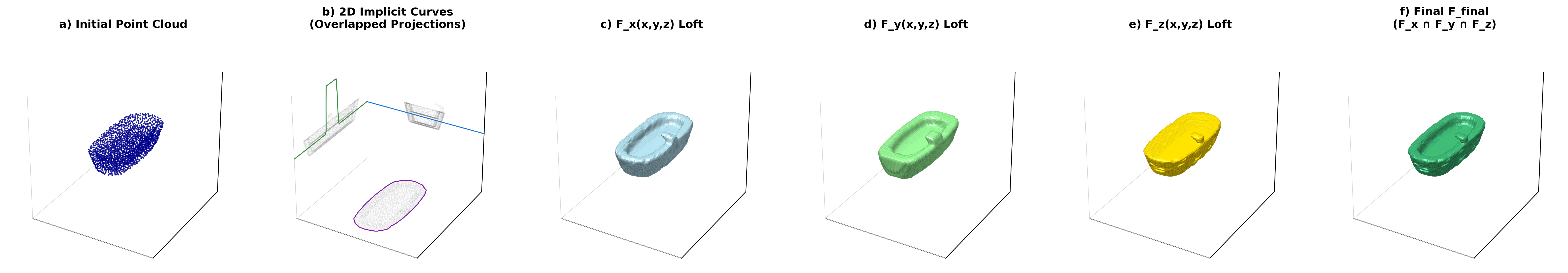} 
\centering \includegraphics[width=\textwidth]{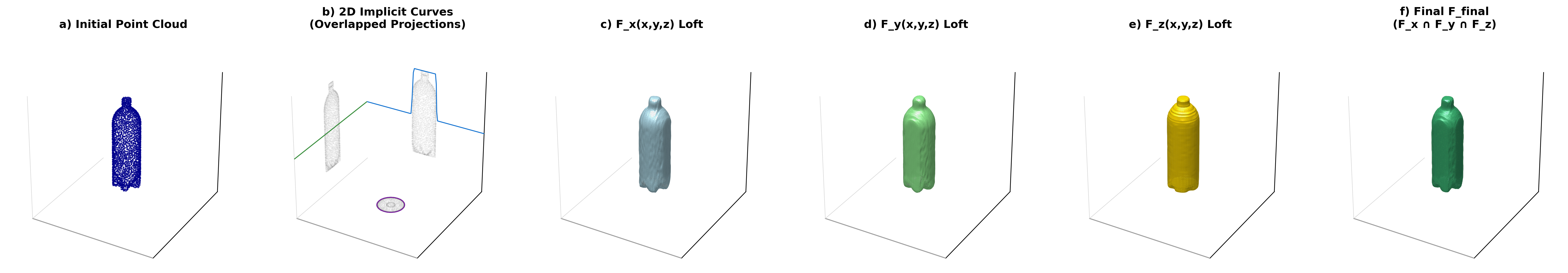} 
\centering \includegraphics[width=\textwidth]{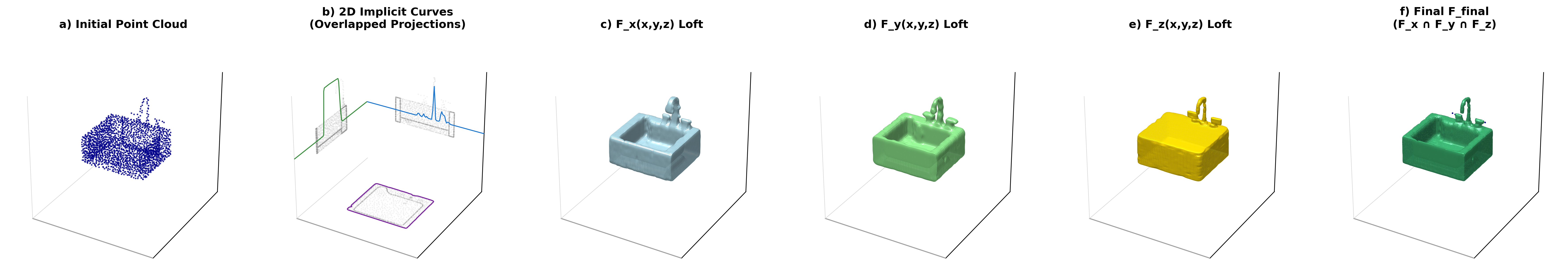} 
\centering \includegraphics[width=\textwidth]{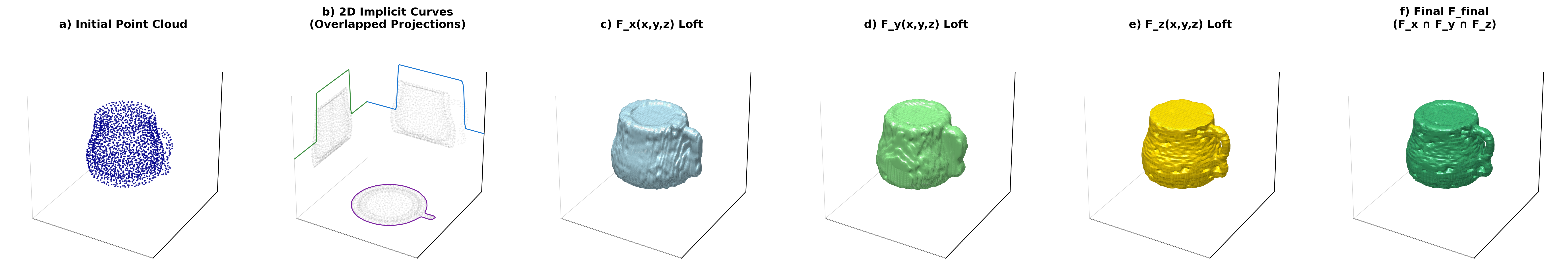} 
\centering \includegraphics[width=\textwidth]{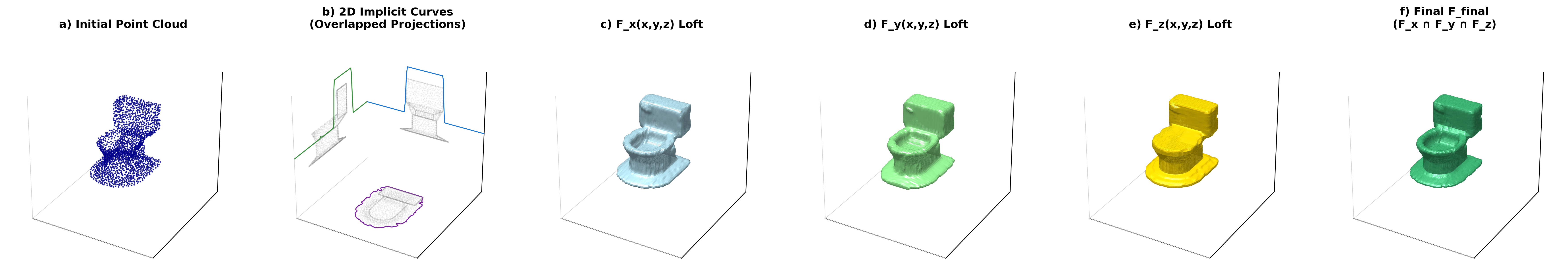} 
\centering \includegraphics[width=\textwidth]{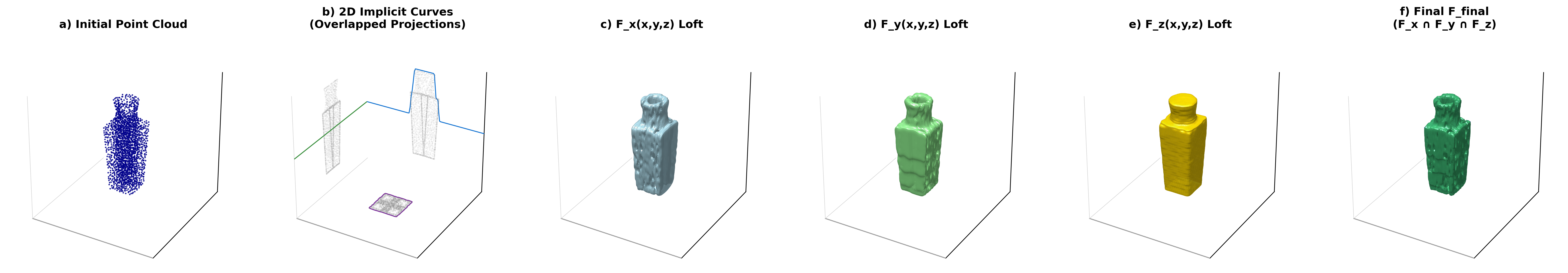}

\caption{ Compact, genus-zero objects are reconstructed reliably from the coarse visual-hull stage alone, with multi-axial refinement primarily improving local shape smoothness and cross-sectional consistency. }
\label{fig:modelnet40_results-1} \end{figure*}

\begin{figure*}[t] 
\centering \includegraphics[width=\textwidth]{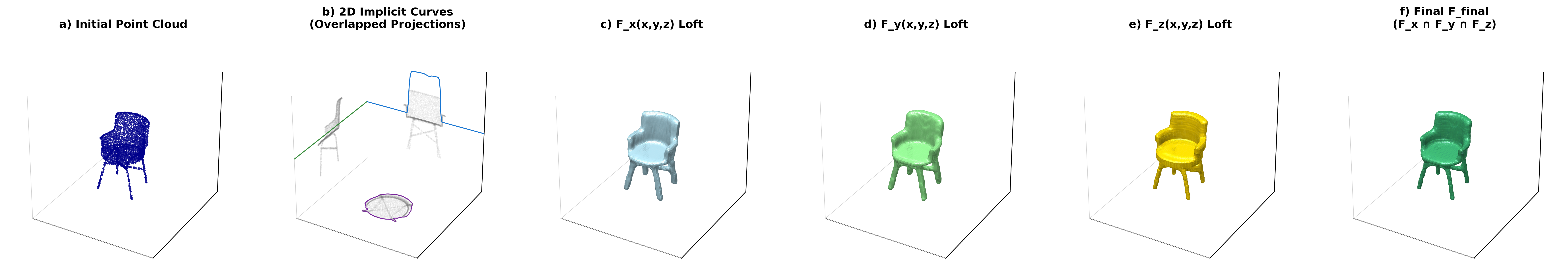} 
\centering \includegraphics[width=\textwidth]{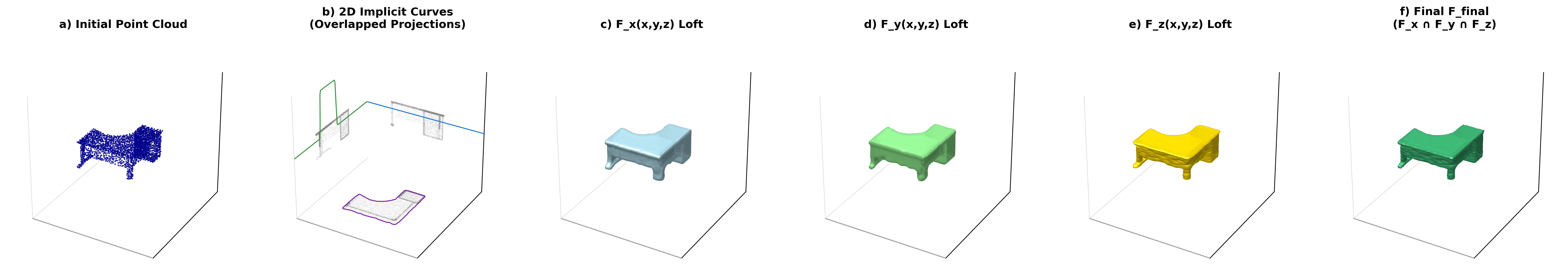} 
\centering \includegraphics[width=\textwidth]{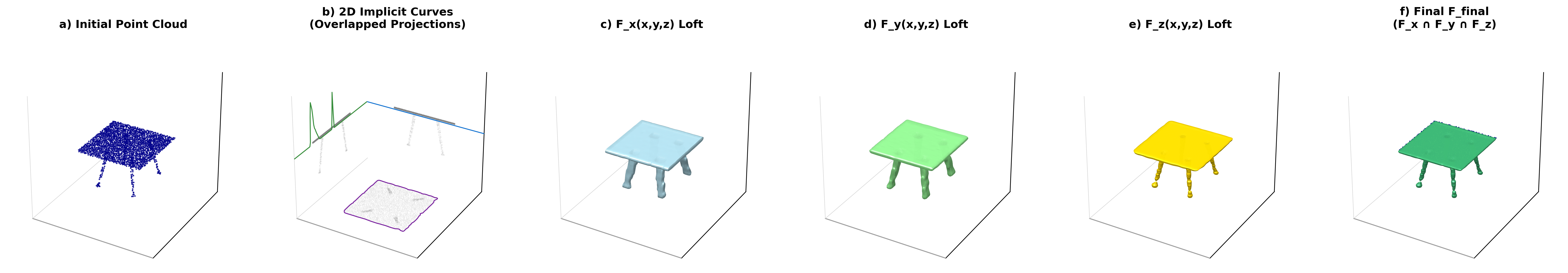} 
\centering \includegraphics[width=\textwidth]{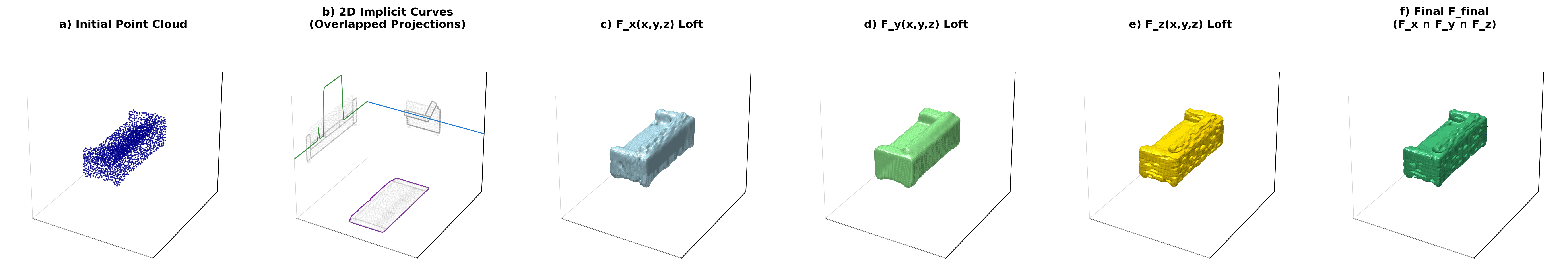} 
\centering \includegraphics[width=\textwidth]{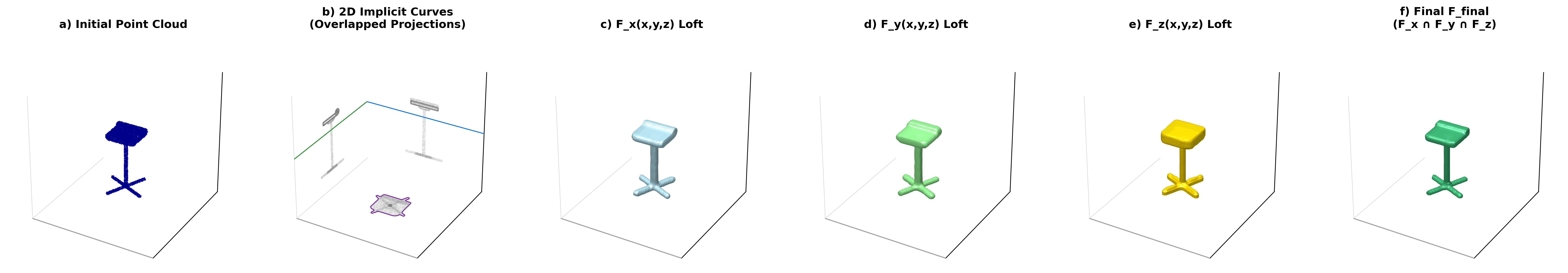}

\caption{ Objects containing supporting structures provide a more demanding test of geometric abstraction due to the presence of thin legs, structural separations, and multiple disconnected components. The results demonstrate the ability of the PSPS reconstruction stage to preserve major structural relationships while maintaining a continuous functional representation. }
\label{fig:modelnet40_results-2} \end{figure*}

\begin{figure*}[t] 

\centering \includegraphics[width=\textwidth]{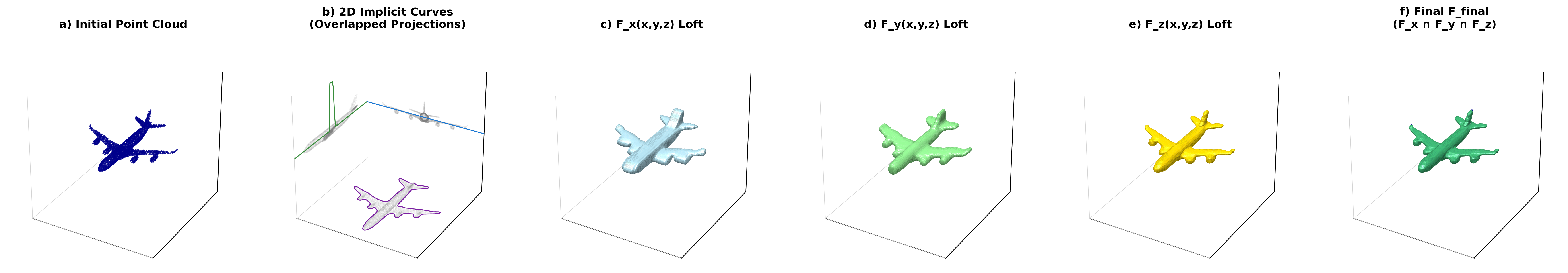} 
\centering \includegraphics[width=\textwidth]{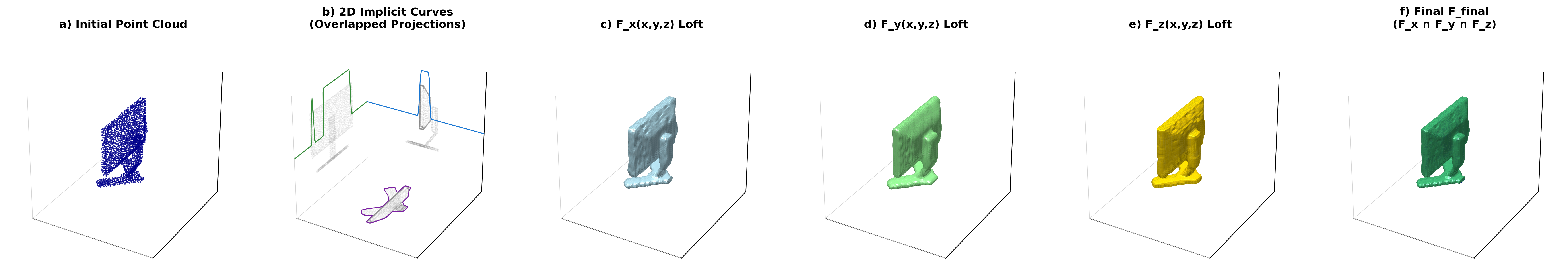} 
\centering \includegraphics[width=\textwidth]{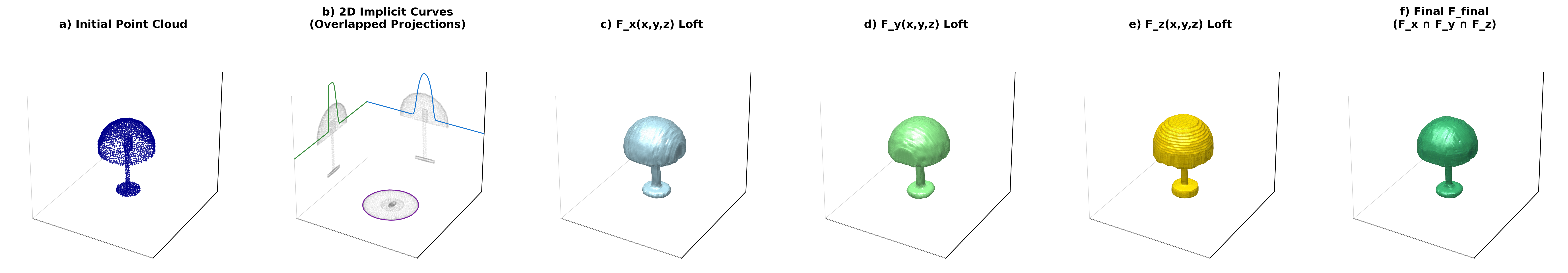} 
\centering \includegraphics[width=\textwidth]{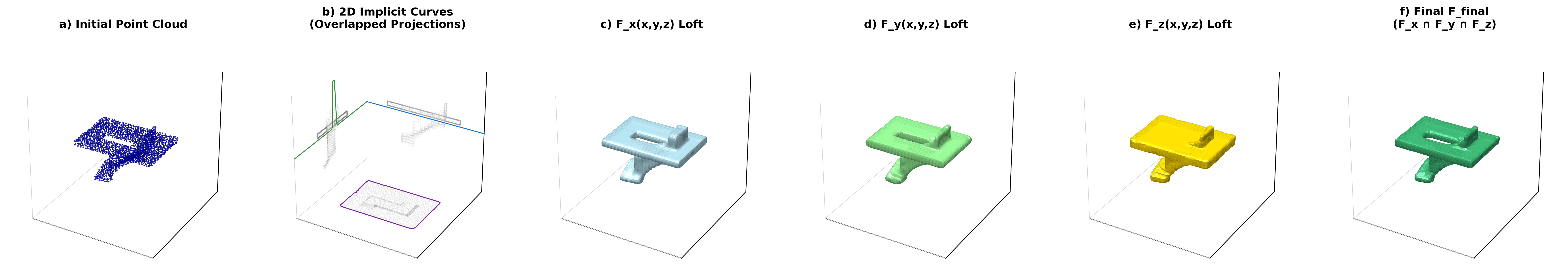} 
\centering \includegraphics[width=\textwidth]{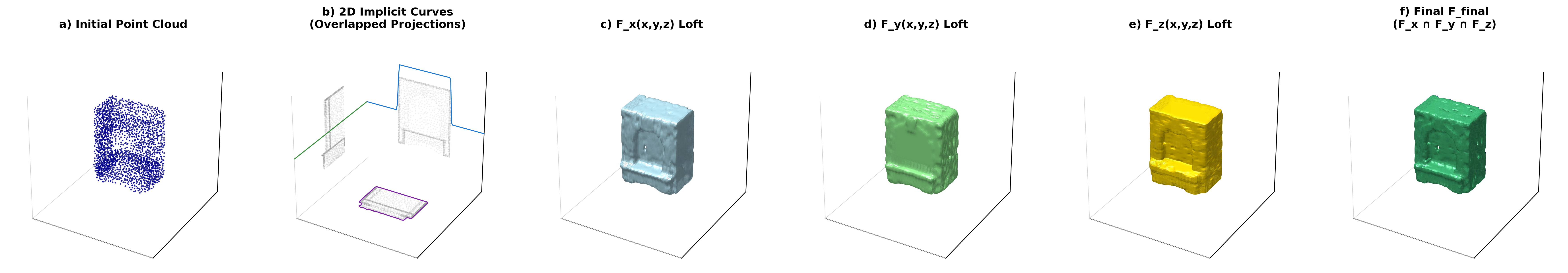} 
\caption{ Engineering objects contain multiple interacting geometric features and often exhibit thin protrusions, local cavities, and partially occluded structures. NeuSOGA3D successfully recovers the dominant topological organisation while preserving characteristic object morphology. }
\label{fig:modelnet40_results-3} \end{figure*}

\begin{figure*}[t] 
\centering \includegraphics[width=\textwidth]{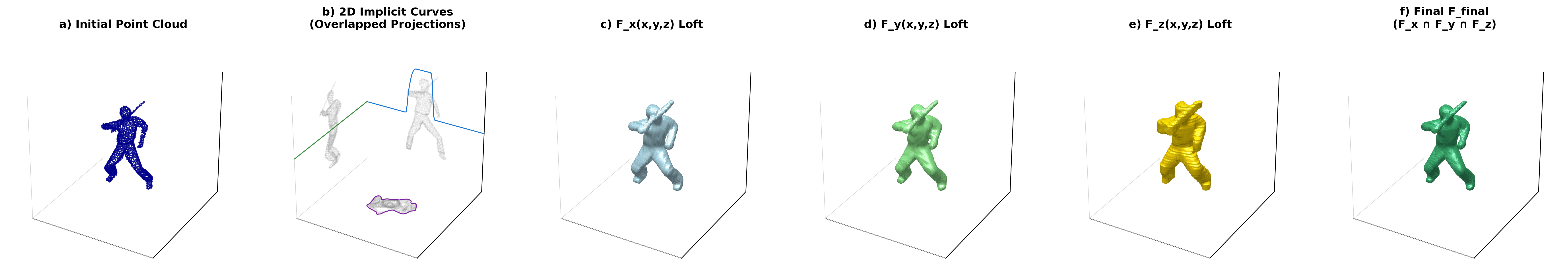} 
\centering \includegraphics[width=\textwidth]{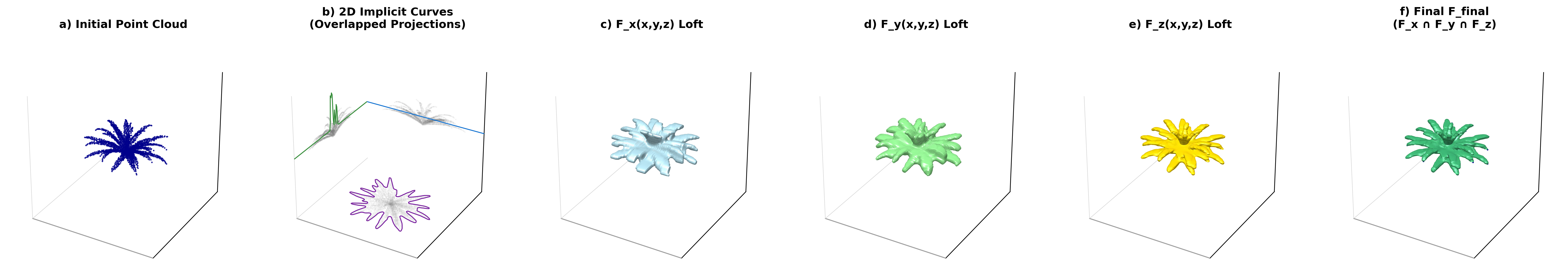} 
\centering \includegraphics[width=\textwidth]{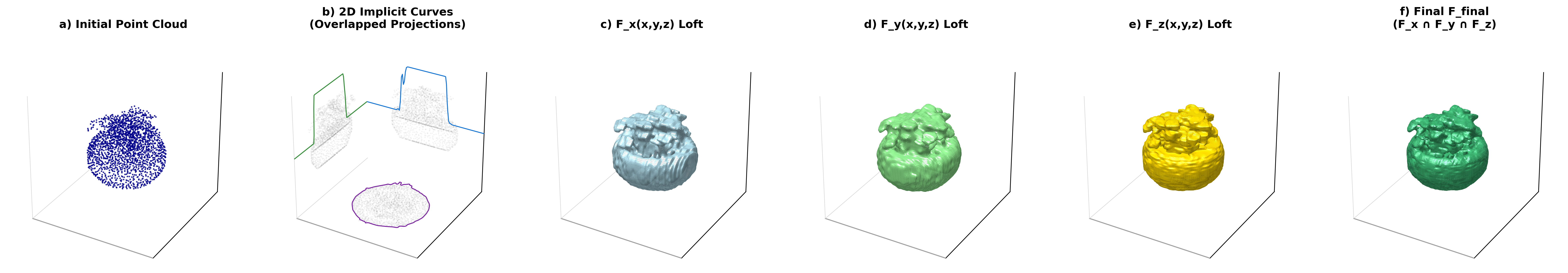} 
\centering \includegraphics[width=\textwidth]{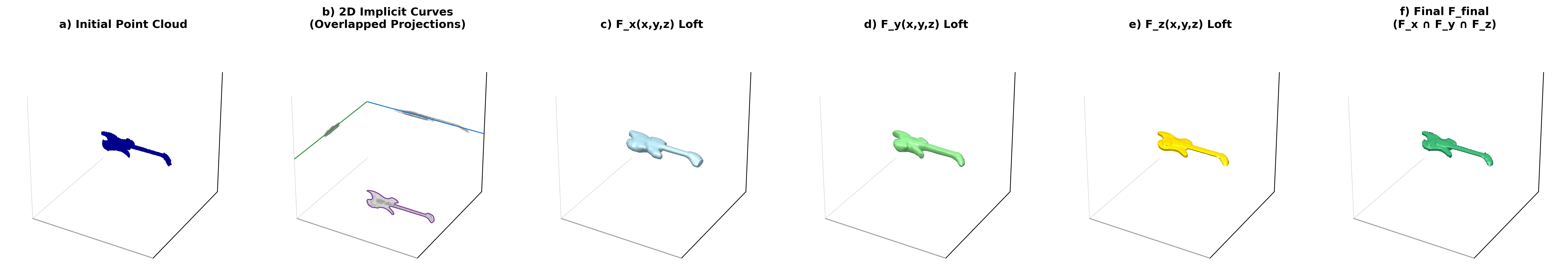} 
\centering \includegraphics[width=\textwidth]{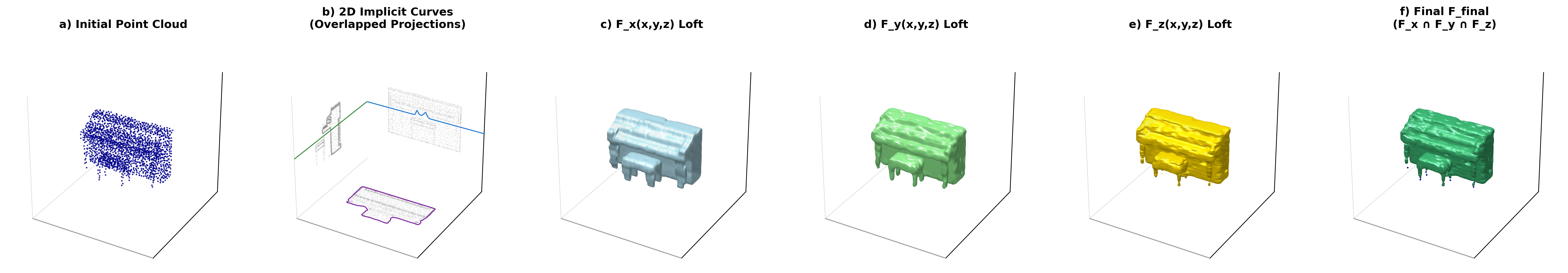} 

\caption{ Articulated objects present greater challenges due to narrow limbs, non-uniform cross sections, and complex local geometry. Although the overall structural organisation remains accurately reconstructed, some loss of high-frequency detail and local feature sharpness can be observed.}
\label{fig:modelnet40_results-4} \end{figure*}

\begin{figure*}[t] 
\centering \includegraphics[width=\textwidth]{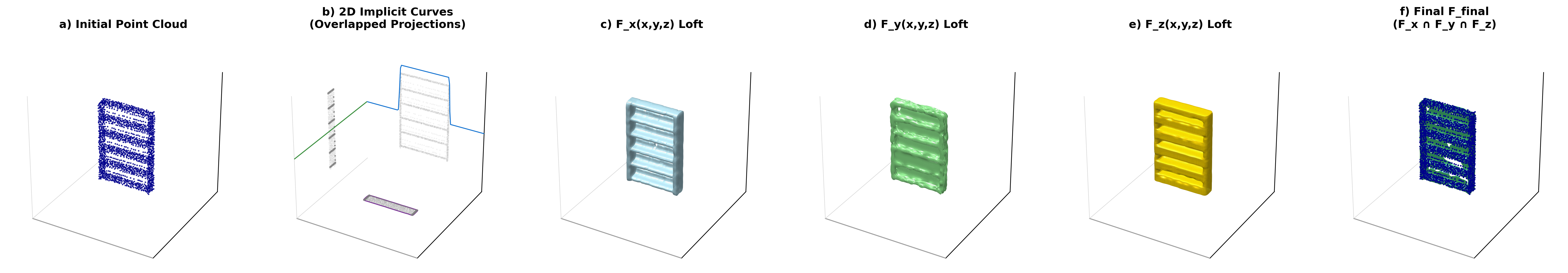} 
\centering \includegraphics[width=\textwidth]{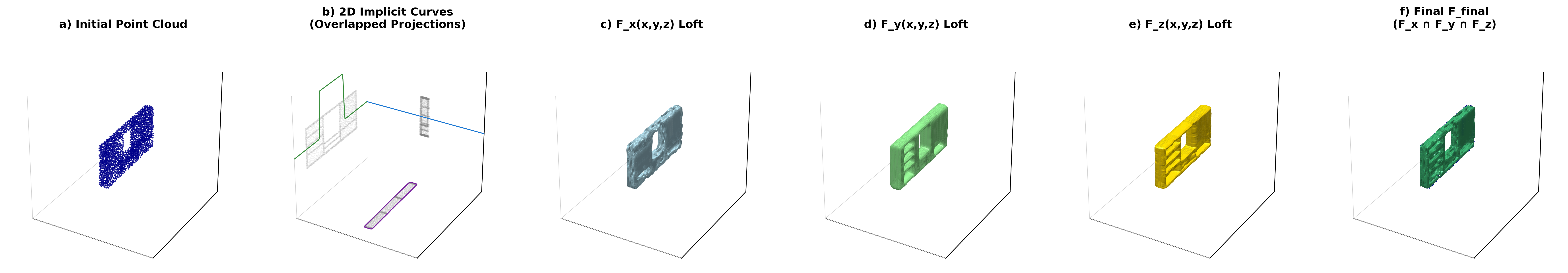} 
\centering \includegraphics[width=\textwidth]{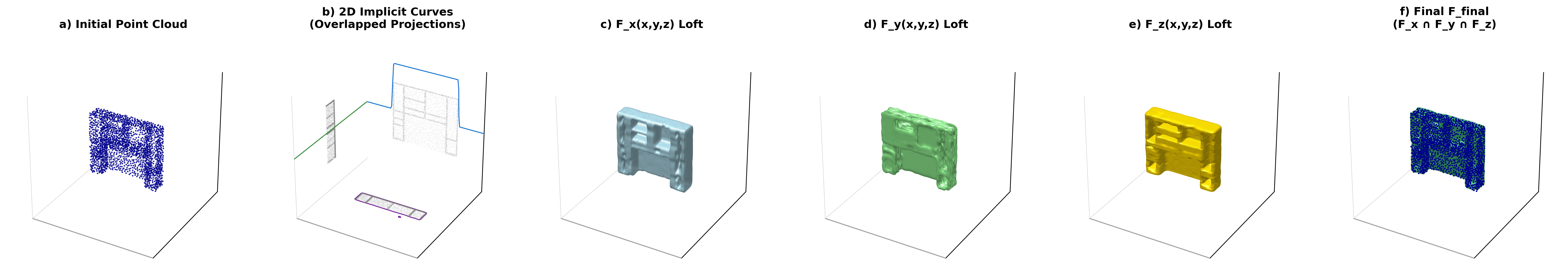} 
\centering \includegraphics[width=\textwidth]{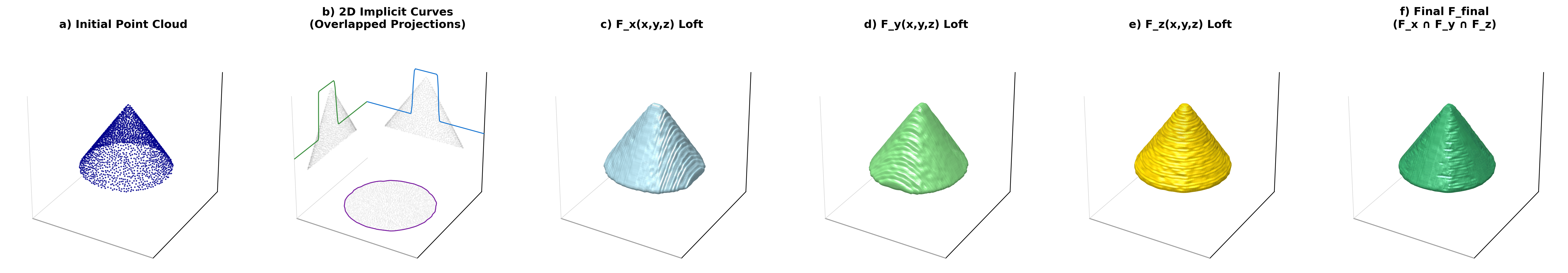} 
\centering \includegraphics[width=\textwidth]{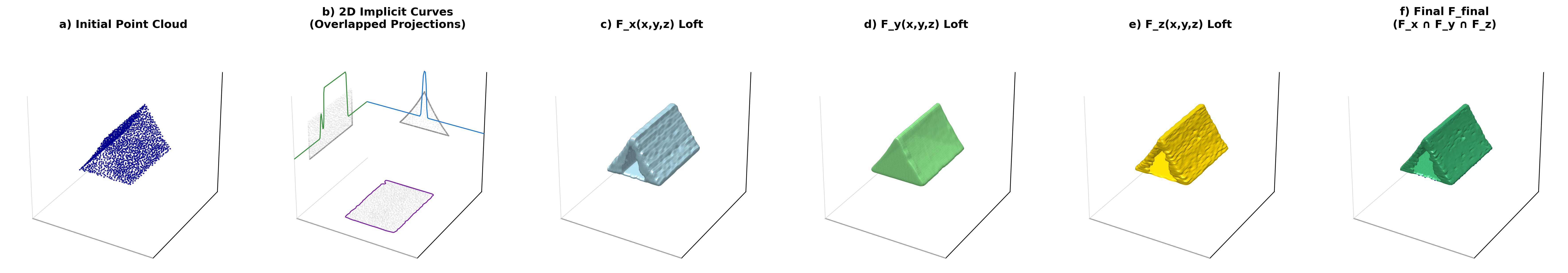} 
\centering \includegraphics[width=\textwidth]{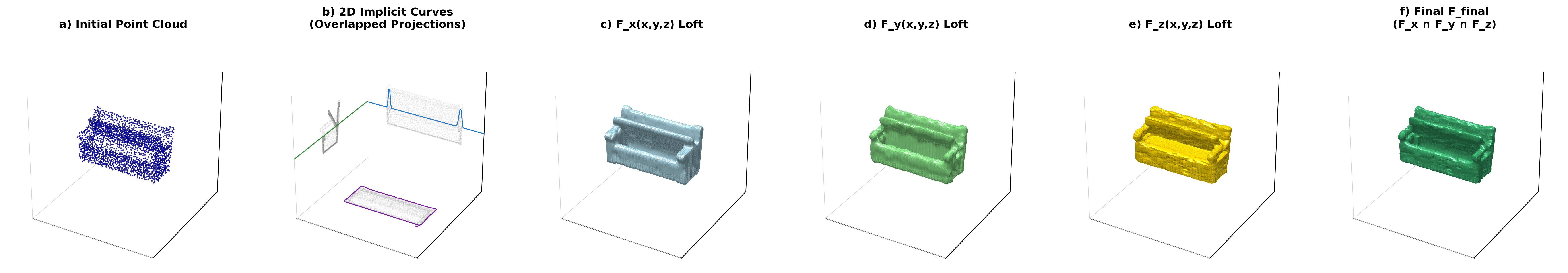}

\caption{ Objects containing multiple compartments, layered structures, or internal voids provide a useful test of topological preservation. The visual results indicate that multi-axial PSPS refinement is able to recover a substantial proportion of the observed structural organisation without requiring latent neural representations. }
\label{fig:modelnet40_results-5} \end{figure*}

Figures~\ref{fig:modelnet40_results-1} to \ref{fig:modelnet40_results-5}  summarizes the reconstruction results. The examples span a wide range of geometric complexity, including compact solids, articulated objects, furniture, engineering components, and structurally complex models containing multiple interacting parts. 

The visual results demonstrate that the proposed framework is capable of generating continuous symbolic geometric representations across a diverse set of object categories without requiring category-specific training, object templates, or learned geometric embeddings.

\subsection{Structural Preservation} The primary purpose of reconstruction within NeuSOGA3D is to recover the underlying structural organisation of an object rather than merely its surface appearance. 

Across the ModelNet40 benchmark, the reconstructed models preserve major geometric relationships including supporting structures, cross-sectional transitions, protrusions, cavities, and inter-component separations. Examples are particularly visible in categories such as chairs, tables, airplanes, bookshelves, and lamps, where preserving structural organisation is more important than reproducing individual surface samples. 

The results indicate that the combination of multi-view symbolic abstraction, PSPS-based volumetric lofting, and shape-preserving constructive solid geometry provides sufficient information to recover meaningful volumetric structure from sparse observations.

\subsection{Explainable Geometric Reconstruction} 

A distinguishing feature of NeuSOGA3D is that every stage of the reconstruction process remains explicit and interpretable. 

Figure~\ref{fig:overview} illustrates the complete reconstruction pipeline. Starting from a raw point cloud, the framework progressively generates symbolic contour representations, implicit spline fields, cross-sectional control slices, PSPS lofts, and finally continuous volumetric geometry. 

Unlike neural implicit representations, where geometry is encoded within distributed network parameters, every reconstructed feature in NeuSOGA3D can be traced directly to identifiable symbolic geometric entities. Consequently, the reconstruction process provides complete geometric provenance from the original observations to the final model. 

The visual examples demonstrate that NeuSOGA3D reconstructs not only surfaces but also interpretable geometric descriptions, thereby supporting the broader objective of explainable geometric intelligence.

\subsection{Symbolic Abstraction and Representation Reuse}

A central hypothesis of this work is that geometric understanding should be achieved through abstraction rather than memorization.

During reconstruction, dense point-cloud observations are transformed into compact collections of symbolic control structures. These control structures capture the essential geometric characteristics of the object while remaining explicit, interpretable, and reusable.

As illustrated in Figure~\ref{fig:overview}, the extracted symbolic representation serves as a common geometric language linking multiple geometric modelling paradigms. The same control structures simultaneously support symbolic implicit reconstruction, constructive solid geometry operations, spline-based modelling, and conventional CAD workflows.

Consequently, NeuSOGA3D does not merely reconstruct a surface. Instead, it reconstructs a reusable geometric representation from which multiple downstream descriptions can be generated without retraining, remeshing, or reverse engineering.

\subsection{Limitations and Future Directions}

The visual results also reveal several limitations of the current implementation. Fine-scale geometric details and sharp structural transitions may be smoothed due to finite cross-sectional sampling and the current $C^1$ continuity constraint imposed during volumetric reconstruction.

Importantly, these limitations arise primarily from the current reconstruction strategy rather than from the symbolic representation itself. Since NeuSOGA3D represents geometry through continuous implicit functions, complex objects can be naturally decomposed into multiple subregions and reconstructed independently before being combined using shape-preserving constructive solid geometry operations.

This hierarchical reconstruction strategy provides a natural pathway towards improved local geometric fidelity while preserving the explainability, symbolic compactness, and representation reuse characteristics of the framework.

\section{Conclusion} 

This paper introduced NeuSOGA3D, a hybrid neuro-symbolic framework for explainable three-dimensional geometric reconstruction from point-cloud observations. Inspired by the interaction between innate structural priors and acquired perceptual experience in human spatial cognition, the framework combines learned perceptual abstraction with explicit symbolic geometric reasoning to reconstruct continuous volumetric geometry. 

The proposed approach integrates three complementary symbolic geometric operators: NeuSOGA implicit spline representations for contour abstraction, Partial Shape-Preserving Splines (PSPS) for volumetric lofting, and shape-preserving constructive solid geometry operators for volumetric fusion. Together, these components provide a unified analytical framework for transforming unorganized point-cloud observations into continuous geometric representations while preserving complete geometric traceability. 

Unlike conventional neural implicit methods, which encode geometry within latent representations, NeuSOGA3D progressively converts observations into explicit symbolic entities including contour abstractions, cross-sectional representations, implicit fields, and control structures. Consequently, every stage of the reconstruction process remains interpretable, editable, and reusable. The resulting representation supports both symbolic geometric reasoning and direct integration into downstream CAD and engineering workflows. 

Experimental results across all forty categories of the ModelNet40 benchmark demonstrate the generality of the proposed framework and its ability to recover structurally meaningful geometric representations across a diverse range of object classes. The visual results further show that the extracted symbolic control structures provide a reusable geometric abstraction from which multiple representations can be generated without retraining, remeshing, or reverse engineering. 

More broadly, this work suggests an alternative perspective on geometric artificial intelligence. Rather than treating three-dimensional reconstruction solely as a problem of numerical approximation, NeuSOGA3D demonstrates how geometric understanding can emerge through symbolic abstraction and structural reasoning. From this viewpoint, the framework represents not merely a reconstruction algorithm, but a step toward explainable geometric intelligence built upon the integration of learned perception and symbolic geometry.

\bibliographystyle{plain}
\bibliography{references}

\end{document}